\documentclass{article} 
\usepackage{iclr2019_conference,times}

\usepackage[letterpaper]{geometry}
\usepackage[parfill]{parskip}

\usepackage{epsfig} 
\usepackage{wrapfig}
\usepackage{xr-hyper,refcount}
\usepackage{csquotes}
\usepackage{natbib}
\usepackage{soul}

\usepackage{algorithm}
\usepackage{algorithmic}

\usepackage[colorlinks]{hyperref}

\usepackage{amsmath, amssymb}
\usepackage{bbm}
\usepackage{url}
\usepackage[english]{babel}
\usepackage{blindtext}
\usepackage{color}
\usepackage{booktabs}
\usepackage{multirow}
\usepackage{xcolor}
\usepackage{subcaption}

\newcommand{\experiments}[1]{\textcolor{blue}{}}

\newcommand{\edc}{\textsc{EDC}}

\newcommand{\prob}{\textsc{ALPF}}

\newcommand{\vct}{\boldsymbol }

\newcommand{\yhat}{\hat{y}}

\newcommand{\answer}{\alpha_{q}}

\usepackage{amsmath,amsfonts,bm}

\def\eqref#1{equation~\ref{#1}}

\def\1{\bm{1}}

\DeclareMathAlphabet{\mathsfit}{\encodingdefault}{\sfdefault}{m}{sl}
\SetMathAlphabet{\mathsfit}{bold}{\encodingdefault}{\sfdefault}{bx}{n}

\DeclareMathOperator*{\argmin}{arg\,min}

\title{Active Learning with Partial Feedback}

\iclrfinalcopy 

\author{
  Peiyun Hu$^{1}$\thanks{This work was done while the author was an intern at Amazon AI},
  Zachary C. Lipton$^{1,3}$,
  Anima Anandkumar$^{2,3}$,
  Deva Ramanan$^1$ \\
  $^{1}$Carnegie Mellon University\\
  $^{2}$California Institute of Technology\\
  $^{3}$Amazon AI\\
  {
    \small
    \href{mailto:peiyunh@cs.cmu.edu}{\nolinkurl{peiyunh@cs.cmu.edu}},
    \href{mailto:zlipton@cmu.edu}{\nolinkurl{zlipton@cmu.edu}},
    \href{mailto:anima@caltech.edu}{\nolinkurl{anima@caltech.edu}},
    \href{mailto:deva@cs.cmu.edu}{\nolinkurl{deva@cs.cmu.edu}}
  }
}

\begin{document}

\maketitle

\begin{abstract}
While many active learning papers 
assume that the learner can simply ask for a label and receive it,
real annotation often presents a mismatch between the form of a label (say, one among many classes),
and the form of an annotation (typically \emph{yes}/\emph{no} binary feedback). 
To annotate examples corpora for multiclass classification,
we might need to ask multiple \emph{yes}/\emph{no} questions,
exploiting a label hierarchy if one is available.
To address this more realistic setting,
we propose active learning with partial feedback (ALPF), 
where the learner must actively choose both 
\emph{which example to label} and 
\emph{which binary question to ask}.
At each step, the learner selects an example,
asking if it belongs to a chosen (possibly composite) class. 
Each answer eliminates some classes,
leaving the learner with a \emph{partial label}.
The learner may then either ask more questions 
about the same example (until an exact label is uncovered) 
or move on immediately, 
leaving the first example partially labeled.
Active learning with partial labels requires 
(i) a sampling strategy to choose (example, class) pairs, 
and (ii) learning from partial labels between rounds.  
Experiments on \emph{Tiny ImageNet} 
demonstrate that our most effective method 
improves 26\% (relative) 
in top-1 classification accuracy 
compared to i.i.d. baselines and standard active learners
given $30\%$ of the annotation budget 
that would be required (naively) to annotate the dataset.
Moreover, ALPF-learners fully annotate TinyImageNet 
at 42\% lower cost. 
Surprisingly, we observe that accounting for per-example annotation costs can alter the conventional wisdom
that active learners should solicit labels for \emph{hard} examples.





\end{abstract}

\section{Introduction}
\vspace{-2px}
\label{sec:intro}
Given a large set of unlabeled images,
and a budget to collect annotations,
how can we learn an accurate image classifier  most economically?
Active Learning (AL) seeks to increase data efficiency
by strategically choosing which examples to annotate.
Typically, AL treats the labeling process as atomic:
every annotation costs the same 
and produces a correct label.
However, large-scale multi-class annotation 
is seldom atomic;
we can't simply ask a crowd-worker to select
one among $1000$ classes if they aren't familiar with our ontology.
Instead, annotation pipelines typically solicit feedback 
through simpler mechanisms such as yes/no questions.
For example, to construct the $1000$-class ImageNet dataset,
researchers first filtered candidates 
for each class via Google Image Search,
then asking crowd-workers questions like
``Is there a Burmese cat in this image?'' \citep{deng2009imagenet}.
For tasks where the Google trick won't work,
we might exploit class hierarchies 
to drill down to the \emph{exact} label.
Costs scale with the number of questions asked.
Thus, real-world annotation costs can 
vary per example \citep{settles2011theories}. 

\begin{wrapfigure}{r}{0.5\textwidth} 
\quad \includegraphics[width=.42\textwidth]{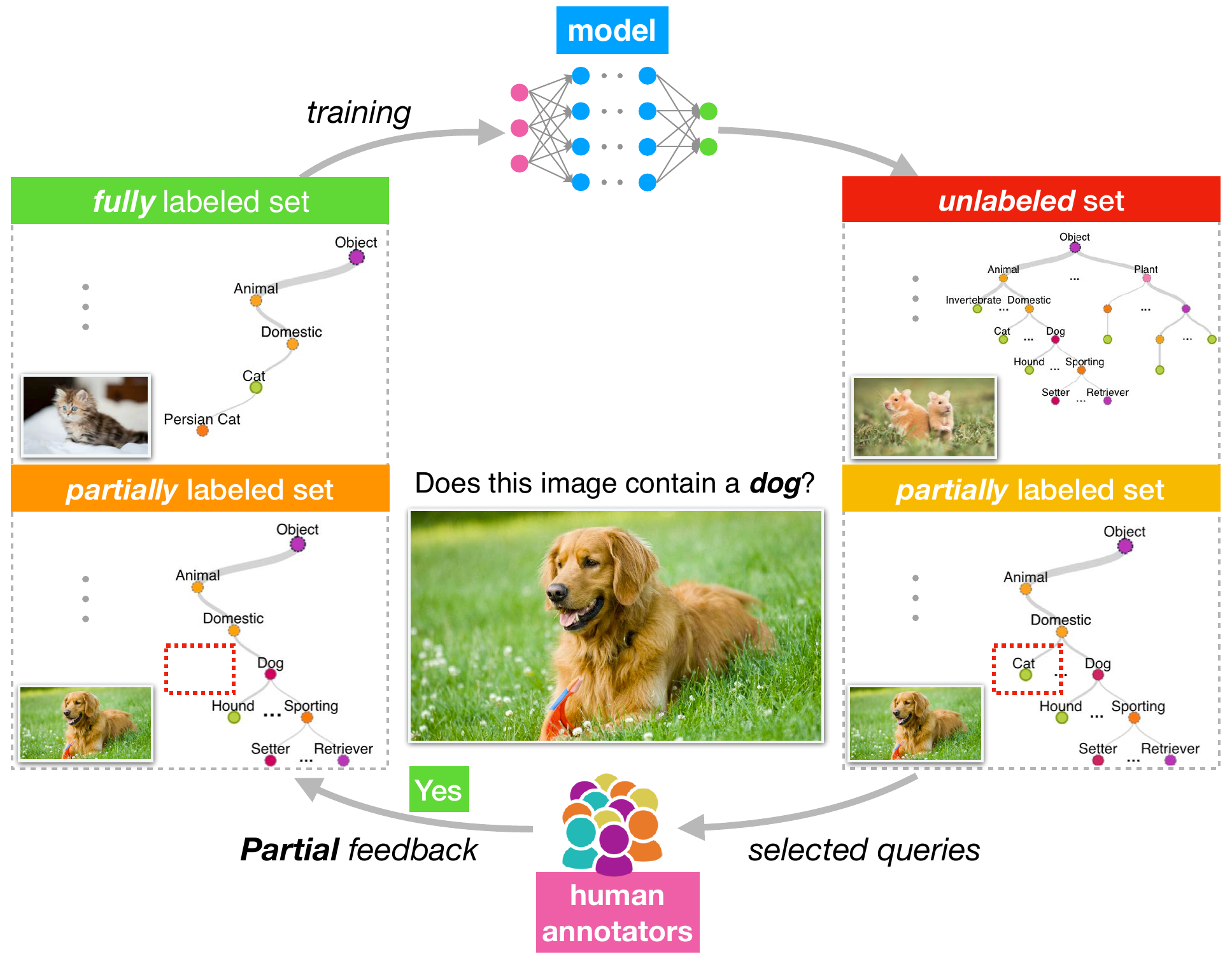}
  \caption{Workflow for an \prob{} learner.}
  \label{fig:framework}
\end{wrapfigure}

\textbf{We propose \emph{Active Learning with Partial Feedback} (\prob{})},
asking, \emph{can we cut costs by actively choosing both  
which examples to annotate,
and which questions to ask?}
Say that for a new image,
our current classifier places 99\% 
of the predicted probability mass 
on various dog breeds.
Why start at the top of the tree -- 
\emph{``is this an artificial object?''} --
when we can cut costs by jumping
straight to dog breeds (Figure \ref{fig:framework})?



\prob{} proceeds as follows: 
In addition to the class labels,
the learner possesses a \emph{pre-defined} collection
of \emph{composite classes},
e.g. \emph{dog} $\supset$ \emph{bulldog}, \emph{mastiff}, ....
At each round, the learner
selects an (example, class) pair.
The annotator responds with binary feedback,
leaving the learner with a \emph{partial label}.
If only the \emph{atomic} class label remains,
the learner has obtained an \emph{exact label}.
For simplicity, we focus on hierarchically-organized collections---trees with atomic classes as leaves
and composite classes as internal nodes.

For this to work, we need a hierarchy of concepts \emph{familiar} to the annotator.
Imagine asking an annotator \emph{``is this a \textbf{foo}?''}
where \emph{foo} represents a category comprised 
of $500$ \emph{random} ImageNet classes.
Determining class membership would be onerous 
for the same reason 
that providing an exact label is: 
It requires the annotator be familiar
with an enormous list of seemingly-unrelated options before answering.
On the other hand, answering \emph{``is this an animal?''}
is easy despite \emph{animal}
being an extremely coarse-grained category ---because
most people already know what an animal is.

We use \emph{active questions} in a few ways. 
To start, in the simplest setup, 
we can select samples at random
but then once each sample is selected, 
choose questions actively until finding the label:
\begin{align*}
&\text{{ML:} ``Is it a dog?''}  &&\text{{Human:} Yes!}\\
&\text{{ML:} ``Is it a poodle?''}  &&\text{{Human:} No!}\\
&\text{{ML:} ``Is it a hound?''}  &&\text{{Human:} Yes!}\\
&\text{{ML:} 
``Is it a Rhodesian 
?''} 
&&\text{{Human:} No!}\\
&\textbf{\color{blue}{{ML:} ``Is it a Dachsund?''}}  &&\textbf{\color{blue}{{Human:} Yes!}}
\end{align*}
In \prob{}, we go one step further. 
Since our goal is to produce accurate classifiers
on tight budget, should we necessarily 
label each example to completion?
After each question, \prob{} learners
have the option of choosing a \emph{different example} 
for the next binary query. 
Efficient learning under \prob{} requires
(i) good strategies for choosing (example, class) pairs,
and (ii) techniques for learning 
from the partially-labeled data 
that results when labeling examples to completion isn't required.

%
%
We first demonstrate an effective scheme
for learning from partial labels.
The predictive distribution is parameterized 
by a softmax over all classes.
On a per-example basis, 
we convert the multiclass problem to a binary classification problem,
where the two classes correspond to the subsets of 
\emph{potential} and \emph{eliminated} classes.
We determine the total probability assigned to \emph{potential} classes
by summing over their softmax probabilities.
For active learning with partial feedback,
we introduce several acquisition functions 
for soliciting partial labels, 
selecting questions among all (example, class) pairs.
One natural method, expected information gain (EIG) 
generalizes the classic maximum entropy heuristic 
to the \prob{} setting. 
Our two other heuristics, EDC and ERC, 
select based on the number of labels 
that we \emph{expect} to see 
\emph{eliminated from} and \emph{remaining in} 
a given partial label, respectively. 

We evaluate \prob{} learners 
on CIFAR10, CIFAR100, and Tiny ImageNet datasets.
In all cases, we use WordNet to impose a hierarchy on our labels.
Each of our experiments simulates rounds of active learning,
starting with a small amount of i.i.d. data 
to warmstart the models,
and proceeding until all examples are exactly labeled. 
We compare models by their test-set accuracy
after various amounts of annotation.
Experiments show that ERC sampling performs best.
On \emph{TinyImageNet}, 
with a budget of 250k binary questions,
ALPF improves in accuracy 
by 26\% (relative) and 8.1\% (absolute)
over the i.i.d. baseline. 
Additionally, ERC \& EDC fully annotate the dataset
with just $491k$ and $484k$ examples binary questions,
respectively (vs 827k), 
a 42\% reduction in annotation cost. 
Surprisingly, we observe
that taking disparate annotation costs into account
may alter the conventional wisdom
that active learners should solicit labels
for \emph{hard} examples.
In \prob{}, \emph{easy} examples might yield less information, but are cheaper to annotate.

\vspace{-2px}

\section{Active Learning with Partial Feedback}
\vspace{-2px}
\label{sec:partial-feedback}



By $\vct x \in \mathcal{R}^d$ and $y \in \mathcal{Y}$ for $\mathcal{Y} = \{\{1\},...,\{k\}\}$,
we denote feature vectors and labels.
Here $d$ is the feature dimension
and $k$ is the number of \emph{atomic} classes.
By \emph{atomic} class,
we mean that they are indivisible.
As in conventional AL,
the agent starts off with an 
unlabeled training set
$\mathcal{D} = \{\vct x_1, ..., \vct x_n\}$.

\textbf{Composite classes }
We also consider a pre-specified collection
of composite classes $\mathcal{C}=\{c_1, ..., c_m\}$,
where each composite class $c_i \subset \{1, ..., k\}$
is a subset of labels such that $|c_i| \geq 1$.
Note that $\mathcal{C}$ includes
both the atomic and composite classes.
In this paper's empirical section,
we generate composite classes
by imposing an existing lexical hierarchy
on the class labels~\citep{miller1995wordnet}.

\textbf{Partial labels }
For an example $i$, we use \emph{partial label}
to describe any element
$\tilde{y}_i\subset\{1,...,k\}$ such that $\tilde{y}_i \supset y_i$.
We call $\tilde{y}_i$ a \emph{partial label}
because it may rule out some classes,
but doesn't fully indicate underlying atomic class.
For example,
$\textit{dog}= \textit{\{akita, beagle, bulldog, ...\}}$
is a valid partial label
when the true label is $\{\textit{bulldog}\}$.
An \prob{} learner eliminates classes,
obtaining successively smaller partial labels,
until only one (the \emph{exact label}) remains.
To simplify notation, in this paper,
by an example's \emph{partial label},
we refer to the smallest partial label available
based on the already-eliminated classes.
At any step $t$ and for any example $i$,
we use $\tilde{y}_i^{(t)}$
to denote the current partial label.
The initial partial label for every example is $\tilde{y}^{0}=\{1,...,k\}$ 
An \emph{exact label} is achieved when the partial label $\tilde{y}_i = y_i$.

\textbf{Partial Feedback }
The set of possible questions
$\mathcal{Q} = \mathcal{X} \times \mathcal{C}$
includes all pairs of examples and composite classes.
An \prob{} learner interacts with annotators
by choosing questions $q\in \mathcal{Q}$.
Informally, we pick a question $q=(\vct x_i, c_j)$
and ask the annotator,
\emph{does $\vct x_i$ contain a $c_j$?}
If the queried example's label belongs to the queried composite class 
($y_i \subset c_j$), 
the answer is 1, else $0$.

Let $\answer$
denote the binary answer
to question $q \in \mathcal{Q}$. 
Based on the partial feedback,
we can compute the new partial label
$\tilde{y}^{(t+1)}$
according to Eq.~\eqref{eq:newlabel},
\begin{align}
  \label{eq:newlabel}
  \tilde{y}^{(t+1)} =
  \left\{
  \begin{array}{ll}
      \tilde{y}^{(t)} \setminus c & \mbox{if } \alpha=0 \\
      \tilde{y}^{(t)} \setminus \overline{c} & \mbox{if } \alpha=1
  \end{array}
  \right.
\end{align}
Note that here $\tilde{y}^{(t)}$ and $c$ are sets, $\alpha$ is a bit, $\overline{c}$ is a set complement,
and that $\tilde{y}^{(t)} \setminus \overline{c}$ and $\tilde{y}^{(t)} \setminus c$ are set subtractions to eliminate 
classes from the partial label based on the answer.

\textbf{Learning Process}
The learning process is simple:
At each round $t$, 
the learner selects
a pair $(\vct x, c)$ for labeling.
Note that a rational agent 
will never select either
(i) an example for which the exact label is known,
or (ii) a pair $(\vct x,c)$
for which the answer is already known, e.g.,
if 
$c \supset \tilde{y}^{(t)}$ or
$c \cap \tilde{y}^{(t)}=\emptyset$.
After receiving binary feedback,
the agent updates the corresponding partial label
$\tilde{y}^{(t)} \rightarrow \tilde{y}^{(t+1)}$,
using Equation \ref{eq:newlabel}.
The agent then re-estimates its model,
using all available non-trivial partial labels
and selects another question $q$.
In batch-mode, the \prob{} learner re-estimates its model
once per $T$ queries
which is necessary when training is expensive (e.g. deep learning).
We summarize the workflow of a \prob{} learner in Algorithm~\ref{alg:alpf}.

\textbf{Objectives }
We state two goals for \prob{} learners.
First, we want to learn predictors with
low error (on exactly labeled i.i.d. holdout data),
given a fixed annotation budget.
Second, we want to fully annotate datasets
at the lowest cost.
In our experiments (Section \ref{sec:experiments}),
a \prob{} strategy dominates on both tasks.

\subsection{Learning from partial labels}
We now address the task of learning a multiclass classifier from partial labels, a fundamental requirement of \prob{},
regardless of the choice of sampling strategy.
At time $t$, our model
$\hat{y}(y, \vct x, \theta^{(t)})$
parameterised by parameters $\theta^{(t)}$
estimates the conditional probability
of an atomic class $y$.
For simplicity, when the context is clear,
we will use
$\vct{\hat{y}}$ to designate the full vector of predicted probabilities over all classes.
The probability assigned to a partial label $\tilde{y}$ can be expressed by marginalizing over the atomic classes that it contains:
$\hat{p}(\tilde{y}^{(t)}, \vct x, \theta^{(t)})
=\sum_{y\in\tilde{y}^{(t)}} \hat{y}(y, \vct x, \theta^{(t)})$.
We optimize our model by minimizing the log loss:
\begin{align}
  \label{eq:loss}
  \mathcal{L}(\theta^{(t)}) =
  - \frac{1}{n} \sum_{i=1}^n \log \left[
  \hat{p}(\tilde{y}_i^{(t)}, \vct x_i, \theta^{(t)})
  \right]
\end{align}
Note that when every example
is exactly labeled,
our loss function simplifies
to the standard cross entropy loss
often used for multi-class classification.
Also note that when every partial label contains the full set of classes,
all partial labels have probability 1 and the update is a no-op.
Finally, if the partial label indicates a composite class such as \emph{dog},
and the predictive probability mass
is exclusively allocated among various breeds of dog, our loss will be $0$.
Models are only updated
when their predictions
disagree (to some degree) with the current partial label.

\subsection{Sampling strategies}

{\bf Expected Information Gain (EIG):}
Per classic uncertainty sampling,
we can quantify a classifer's uncertainty 
via the entropy of the predictive distribution.
In AL, each query returns an exact label, 
and thus the post-query entropy is always $0$. 
In our case, each answer to the query 
yields a different partial label.
We use the notation $\vct \yhat_0 $,
and $\vct \yhat_1$
to denote consequent predictive distributions for each answer (no or yes).
We generalize maximum entropy to \prob{}
by selecting questions with greatest
\emph{expected reduction in entropy}.
\begin{equation}
  \label{eq:eig}
  EIG_{(\vct x, c)} = S(\vct \yhat)
  - \left[ \hat{p}(c, \vct x, \theta)
   S(\vct \yhat_1)\\
  + (1 - \hat{p}(c, \vct x, \theta))
   S(\vct \yhat_0) \right]
\end{equation}
where $S(\cdot)$ is the entropy function. 
It's easy to prove that EIG is maximized when
$\hat{p}(c, \vct x, \theta)=0.5$.

\begin{figure}
\centering
    \begin{minipage}[t][][t]{.48\linewidth}
    \begin{algorithm}[H]
       \caption{Active Learning with Partial Feedback}
       \label{alg:alpf}
       \begin{algorithmic}
         \STATE {\bfseries Input:} $\mathbf{X} \leftarrow (\mathbf{x}_{1}, \dots, \mathbf{x}_{N})$, $\mathbf{Q} \leftarrow (\mathbf{q}_{1}, \dots, \mathbf{q}_{M})$, $K$, $T$.
         \STATE {\bfseries Input:} $\mathcal{D} \leftarrow [\vct x_i]_{i=1}^{N}$, $\mathcal{C} \leftarrow [c_{j}]_{j=1}^{M}$, $k$, $T$
         \STATE {\bfseries Initialize:} $\tilde{y}_i^{(0)} \leftarrow \{1,\dots,k\}$, $\theta \leftarrow \theta^{(0)}$, $t \leftarrow 0$
         \REPEAT	
         \STATE Score every $(\vct x_{i}, c_{j})$ with $\theta$
         \REPEAT
         \STATE Select $(\vct x_{i^{*}}, c_{j^{*}})$ with the best score
         \STATE Query $c_{j^*}$ on data $\vct x_{i^*}$ 
         \STATE Receive feedback $\alpha$
         \STATE Update $\tilde{y}^{(t+1)}_{i^{*}}$ according to $\alpha$ 
        \STATE $t \leftarrow t+1$
         \UNTIL{($t~\text{mod}~T = 0$) or ($\forall i, |\tilde{y}^{(t)}_{i}|=1$)}
         \STATE $\theta \leftarrow \argmin_{\theta} \mathcal{L}(\theta)$
         \UNTIL{
           $\forall i, |\tilde{y}^{(t)}_{i}|=1$ or $t$ exhausts budget
           }
       \end{algorithmic}
     \end{algorithm}
\end{minipage}
\hspace{.02\linewidth}~\begin{minipage}[t][][t]{.48\linewidth}
     \begin{table}[H]
      \caption{Learning from partial labels on Tiny ImageNet. These results demonstrate the usefulness of our training scheme absent the additional complications due to ALPF. 
In each row, $\gamma \%$ of examples
are assigned labels at the \emph{atomic class} (Level $0$).
Levels 1, 2, and 4 denote 
progressively coarser composite labels
tracing through the WordNet hierarchy.
    }
    \vspace{2px}
      \label{tab:loss}
      \centering
      \begin{tabular}{c c c c c}
        \toprule
        \multirow{2}{*}{$\gamma$(\%)} & $\gamma$ & \multicolumn{3}{c}{$(1-\gamma)$} \\
        \cmidrule{3-5}
                                     &Level $0$ &Level $1$ & Level $2$ & Level $4$\\
        \midrule
        20 & 0.285 & \textbf{+0.113} & +0.086 & +0.025 \\
        40 & 0.351 & +0.079 & +0.056 & +0.016 \\
        60 & 0.391 & +0.051 & +0.036 & +0.018 \\
        80 & 0.432 & +0.015 & +0.017 & -0.009 \\
        \midrule
        100 & 0.441 & - & - & - \\
        \bottomrule
      \end{tabular}
    \end{table}
 \end{minipage}
 \vspace{-5px}
\end{figure}

{\bf Expected Remaining Classes (ERC):}
Next, we propose ERC, a heuristic that suggests arriving as quickly as possible at exactly-labeled examples.
At each round, ERC selects those examples for which
the expected number of remaining classes is fewest:
\begin{align}
  \label{eq:erc}
  ERC_{(\vct x, c)} = \hat{p}(c, \vct x, \theta)
  || \vct \yhat_1 ||_0
  +(1-\hat{p}(c, \vct x, \theta))  ||\vct \yhat_0||_0,
\end{align}
where $|| \vct \yhat_\alpha ||$
is the size of the partial label following given answer $\alpha$.
ERC is minimized when
the result of the feedback will produce an exact label with probability $1$.
For a given example $\vct x_i$, 
if $||\vct \yhat_i||_0 = 2$
containing  only the potential classes
(e.g.) \emph{dog} and \emph{cat},
then with certainty, 
ERC will produce an exact label by querying the class \{\emph{dog}\} (or equivalently \{\emph{cat}\}).
This heuristic is inspired by \cite{cour2011learning},
which shows that the partial classification loss
(what we optimize with partial labels)
is an upper bound of the true classification loss
(as if true labels are available) with a linear factor of $\frac{1}{1-\varepsilon}$,
where $\varepsilon$ is ambiguity degree and $\varepsilon \propto |\tilde{y}|$.
By selecting $q \in \mathcal{Q}$ 
that leads to the smallest $|\tilde{y}|$,
we can tighten the bound to make optimization with partial labels more effective.

{\bf Expected Decrease in Classes (EDC):}
More in keeping with the traditional goal of minimizing uncertainty, we might choose \edc,
the sampling strategy which we expect to result in the greatest reduction in the number of potential classes.
We can express EDC as the difference
between the number of potential labels (known)
and the expected number of potential labels remaining: $EDC_{(\vct x, c)} = |\tilde{y}^{(t)}| - ERC_{(\vct x, c)}$.





\vspace{-2px}

\section{Experiments}
\vspace{-2px}
\label{sec:experiments}


We evaluate \prob{} algorithms
on the CIFAR10, CIFAR100, and Tiny ImageNet datasets,
with training sets of 50k, 50k, and 100k examples,
and 10, 100, and 200 classes respectively.
After imposing the Wordnet hierarchy on the label names,
the size of the set of possible binary questions $|\mathcal{C}|$
for each dataset are $27$, $261$, and $304$, respectively.
The number of binary questions between re-trainings
are 5k, 15k, and 30k, respectively.
By default, we warm-start each learner
with the same 5\% of training examples
selected i.i.d. and exactly labeled.
Warm-starting has proven essential
in other papers combining deep and active learning \citep{shen2018deep}.
Our own analysis (Section \ref{sec:diagnostic})
confirms the importance of warm-starting although the affect appears variable across acquisition strategies.


\textbf{Model }
For each experiment,
we adopt the widely-popular ResNet-18 architecture~\citep{he2016deep}.
Because we are focused on active learning
and thus seek fundamental understanding
of this new problem formulation,
we do not complicate the picture with any
fine-tuning techniques.
Note that some leaderboard scores circulating on the Internet appear to have far superior numbers. 
This owes to pre-training on the full ImageNet dataset (from which Tiny-ImageNet was subsampled and downsampled), constituting a target leak. 

We initialize weights with the \emph{Xavier} technique \citep{glorot2010understanding}
and minimize our loss using the Adam~\citep{kingma2014adam} optimizer,
finding that it outperforms SGD
significantly when learning from partial labels.
We use the same learning rate of $0.001$
for all experiments,
first-order momentum decay ($\beta_1$) of $0.9$,
and second-order momentum decay ($\beta_2$) of $0.999$.
Finally, we train with mini-batches of $200$ examples
and perform standard data augmentation techniques
including random cropping, resizing,
and mirror-flipping.
We implement all models in MXNet
and have posted our code publicly\footnote{{\small Our implementations of \prob{} learners are available at: \url{https://github.com/peiyunh/alpf}}}.

\textbf{Re-training }
Ideally, we might update models
after each query, but 
this is too costly.
Instead, following \cite{shen2018deep} and others,
we alternately query labels and update our models in rounds.
We warm-start all experiments with 5\% labeled data 
and iterate until every example is exactly labeled.
At each round, we re-train our classifier from scratch with random initialization.
While we could initialize the new classifier
with the previous best one (as in \cite{shen2018deep}),
preliminary experiments showed
that this faster convergence
comes at the cost of worse performance,
perhaps owing to severe over-fitting
to labels acquired early in training.
In all experiments, for simplicity,
we terminate the optimization after 75 epochs.
%
Since $30k$ questions per re-training (for TinyImagenet)
seems infrequent,
we compared against 10x more frequent re-training
More frequent training conferred no benefit (Appendix \ref{sec:retrain}).

\subsection{Learning from partial labels}


Since the success of \prob{} depends in part on learning from partial labels,
we first demonstrate the efficacy of learning from partial labels with our loss function
when the partial labels are given a priori.
In these experiments we simulate a partially labeled dataset and show that the learner achieves significantly better accuracy
when learning from partial labels than if it excluded the partial labels
and focused only on  exactly annotated examples.
Using our WordNet-derived hierarchy,
we conduct experiments with partial labels at different levels of granularity.
Using partial labels from one level above the leaf, {\it German shepherd} becomes {\it dog}.
Going up two levels, it becomes {\it animal}.

We first train a standard multi-class classifier with $\gamma$~(\%) {\it exactly labeled} training data and then another classifier
with the remaining $(1-\gamma)\%$ {\it partially labeled}
at a different granularity (level of hierarchy).
We compare the classifier performance
on holdout data both \emph{with} and \emph{without} adding {\it partial labels}
in Table~\ref{tab:loss}.
We make two key observations:
(i) additional coarse-grained partial labels
improve model accuracy
(ii) as expected, the improvement diminishes
as partial label gets coarser.
These observations suggest
we can learn effectively given a mix of exact and partial labels.

\subsection{Sampling strategies}


{\bf Baseline }
This learner samples examples at random.
Once an example is sampled,
the learner applies top-down binary splitting---choosing 
the question that most evenly splits the probability mass, 
see Related Work for details---
with a uniform prior over the classes
until that example is exactly labeled.

{\bf AL } To disentangle the effect of active sampling of questions and samples, we compare to conventional AL approaches selecting examples with uncertainty sampling but selecting questions as baseline.
%

{\bf AQ }
\emph{Active questions} learners,
choose examples at random
but use partial feedback strategies
to efficiently label those examples, 
moving on to the next example
after finding an example's exact label.

{\bf \prob}
ALPF learners are free to choose any (example, question) pair at each turn,
Thus, unlike AL and AQ, \prob{} learners
commonly encounter partial labels during training.

\begin{figure}
  \centering
  \includegraphics[width=\linewidth]{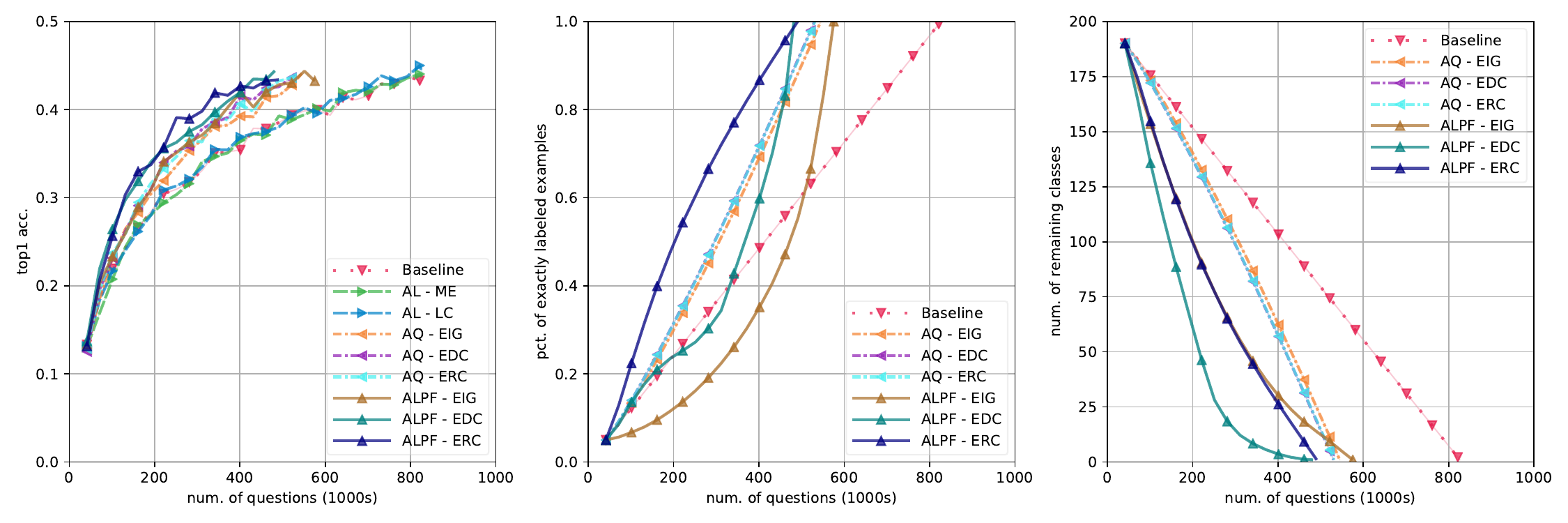}

  \caption{The progression of top1 classification accuracy (left), percentage of exactly labeled training examples (middle), and average number of remaining classes (right).}
  \label{fig:main}
  \vspace{-10px}
\end{figure}

{\bf Results}
We run all experiments
until fully annotating the training set.
We then evaluate each method
from two perspectives:
classification and annotation.
We measure each classifiers' top-1 accuracy
at each annotation budget.
To quantify annotation performance,
we count the number questions
required to {\it exactly label}
all training examples.
We compile our results in Table~\ref{tab:main},
rounding costs to 10\%, 20\% etc.
The budget includes the (5\%) i.i.d. data for  warm-starting.
%
Some key results:
(i) vanilla active learning does not improve
over i.i.d. baselines,
confirming similar observations
on image classification by \cite{sener2017active};
%
%
(ii) AQ provides a dramatic improvement
over baseline. 
The advantage persists throughout training.
These learners sample examples randomly and label to completion (until an exact label is produced) before moving on, 
differing only in how efficiently they annotate data.
(iii) On Tiny ImageNet, at 30\% of budget,
\prob{}-ERC outperforms AQ methods by 4.5\%
and outperforms the i.i.d. baseline by 8.1\%.

\begin{table}[t]
\small
  \caption{Results on Tiny ImageNet (N/A indicates data has been fully labeled)}
  \label{tab:main}
  \centering
  \begin{tabular}{l|cccccc|c}
    \toprule
    & \multicolumn{6}{c|}{\textbf{Annotation Budget}} & \textbf{Labeling Cost} \\
    & \multicolumn{6}{c|}{(w.r.t. baseline labeling cost)} &  \\
    \midrule
    & 10\% & 20\% & 30\% & 40\% & 50\% & 100\% & \\
    \midrule
    \multicolumn{8}{c}{\textbf{TinyImageNet}}\\
    \midrule
    Baseline & 0.186 & 0.266 & 0.310 & 0.351 & 0.354 & 0.441 & 827k \\
    \midrule
    AL - ME & 0.169 & 0.269 & 0.303 & 0.347 & 0.365 & - & 827k \\
    AL - LC & 0.184 & 0.262 & 0.313 & 0.355 & 0.369 & - & 827k \\
    \midrule
    AQ - EIG & 0.186 & 0.283 & 0.336 & 0.381 & 0.393 & - & 545k \\
    AQ - EDC & 0.196 & 0.291 & 0.353 & 0.386 & 0.415 & - & 530k \\
    AQ - ERC & 0.194 & 0.295 & 0.346 & 0.394 & 0.406 & - & 531k \\
    \midrule
    ALPF - EIG & 0.203 & 0.289 & 0.351 & 0.384 & 0.420 & - & 575k \\
    ALPF - EDC & \textbf{0.220} & 0.319 & 0.363 & 0.397 & 0.420 & - & 482k \\
    ALPF - ERC & 0.207 & \textbf{0.330} & \textbf{0.391} & \textbf{0.419} & \textbf{0.427} & - & 491k \\
    \midrule
    \multicolumn{8}{c}{\textbf{CIFAR100}} \\
    \midrule
    Baseline & 0.252 & 0.340 & 0.412 & 0.437 & 0.469 & 0.537 & 337k \\
    \midrule
    AL - ME & 0.237 & 0.321 & 0.388 & 0.419 & 0.458 & - & 337k \\
    AL - LC & 0.247 & 0.332 & 0.398 & 0.432 & 0.468 & - & 337k \\
    \midrule
    AQ - EIG & 0.266 & 0.354 & 0.443 & 0.485 & 0.502 & - & 208k \\
    AQ - EDC & 0.264 & 0.366 & 0.439 & 0.483 & 0.508 & - & 215k \\
    AQ - ERC & 0.256 & 0.366 & 0.453 & 0.479 & 0.496 & - & 215k \\
    \midrule
    ALPF - EIG & 0.263 & 0.341 & 0.423 & 0.466 & 0.497 & - & 235k \\
    ALPF - EDC & \textbf{0.281} & 0.367 & 0.442 & 0.479 & 0.518 & - & 193k \\
    ALPF - ERC & 0.273 & \textbf{0.379} & \textbf{0.464} & \textbf{0.502} & \textbf{0.526} & - & 187k \\
    \midrule
    \multicolumn{8}{c}{\textbf{CIFAR10}}\\
    \midrule
    Baseline  & 0.645 & 0.718 & 0.757 & 0.778 & 0.792 & 0.829 & 170k \\
    \midrule
    AL - ME & 0.663 & 0.709 & 0.759 & 0.763 & 0.800 & - & 170k \\
    AL - LC & 0.644 & 0.724 & 0.753 & 0.780 & 0.792 & - & 170k \\
    \midrule
	AQ - EIG & 0.654 & 0.747 & 0.791 & 0.806 & 0.823 & - & 89k \\
    AQ - EDC & 0.675 & 0.746 & 0.784 & 0.789 & 0.826 & - & 95k \\
    AQ - ERC & 0.682 & 0.750 & 0.771 & 0.811 & 0.822 & - & 96k  \\
    \midrule
    ALPF - EIG & 0.673 & 0.741 & 0.786 & 0.815 & 0.813 & - & 124k \\
    ALPF - EDC & \textbf{0.676} & \textbf{0.752} & \textbf{0.797} & 0.832 & N/A & -     & 74k \\
    ALPF - ERC & 0.670 & 0.743 & \textbf{0.797} & \textbf{0.833} & N/A & - & 74k \\
    \bottomrule
  \end{tabular}
\end{table}


\subsection{Diagnostic analyses}
\label{sec:diagnostic}
First, we study how different
amounts of warm-starting affects
\prob{} learners' performance with a small set of i.i.d. labels.
Second, we compare the selections due to ERC and EDC
to those produced through uncertainty sampling.
Third, we note that while EDC and ERC
appear to perform best on our problems,
they may be vulnerable to excessively focusing
on classes that are trivial to recognize.
We examine this setting via an adversarial dataset intended to break the heuristics.

{\bf Warm-starting }
We compare the performance of each strategy under
different percentages (0\%, 5\%, and 10\%)
of pre-labeled i.i.d. data (Figure \ref{fig:warmstart}, Appendix \ref{sec:warmstart}).
Results show that ERC works properly
even without warm-starting,
while EIG benefits from a 5\% warm-start
and EDC suffers badly without warm-starting.
We observe that 10\% warm-starting yields no further improvement.

\begin{figure}
    \centering
  \begin{minipage}{0.28\linewidth}
    \centering
  \includegraphics[width=1.0\linewidth]{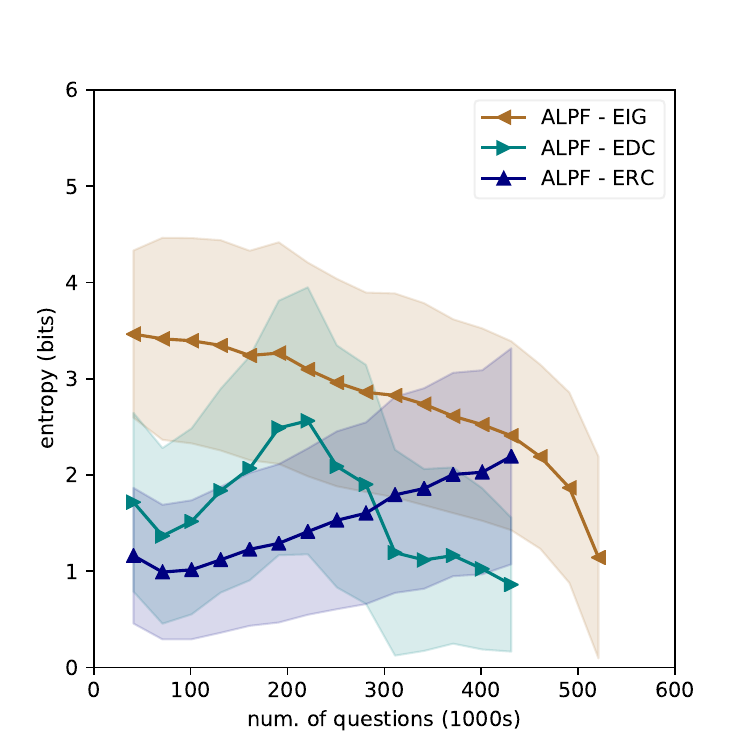}
  \caption{Classifier confidence (entropy of softmax layer) on \emph{selected} examples.
  }
  \label{fig:entropy}
  \end{minipage}
  \centering
  \hspace{.02\linewidth}
  \begin{minipage}{0.668\linewidth}
  \begin{subfigure}{.48\textwidth}
	\centering
    \includegraphics[width=\linewidth]{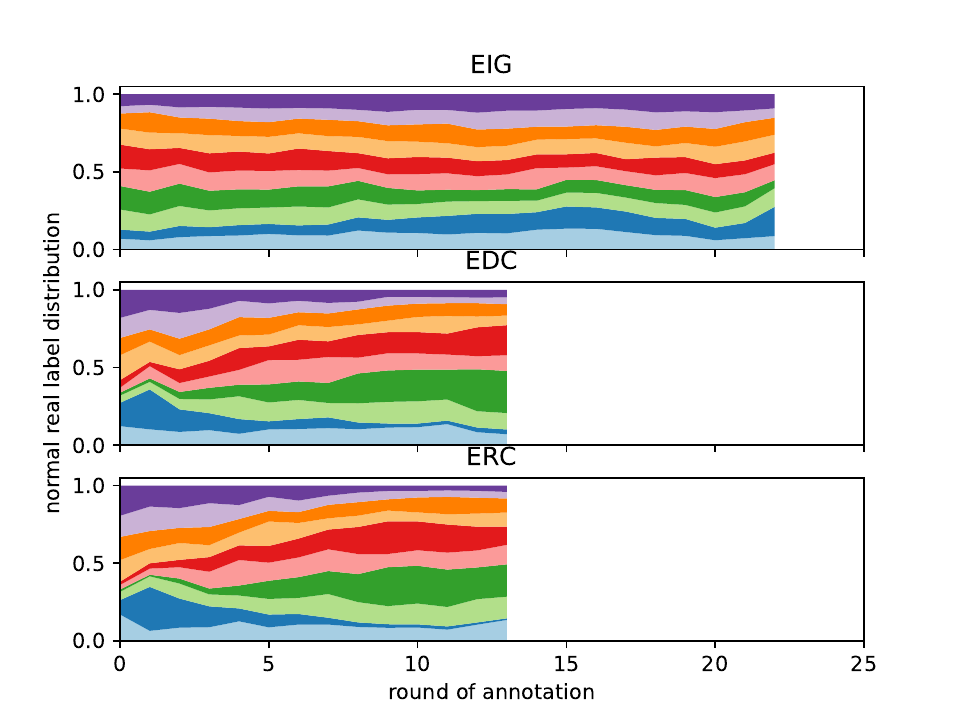}
    \caption{Normal CIFAR10}
  \end{subfigure}
  \begin{subfigure}{.48\textwidth}
    \centering
	\includegraphics[width=\linewidth]{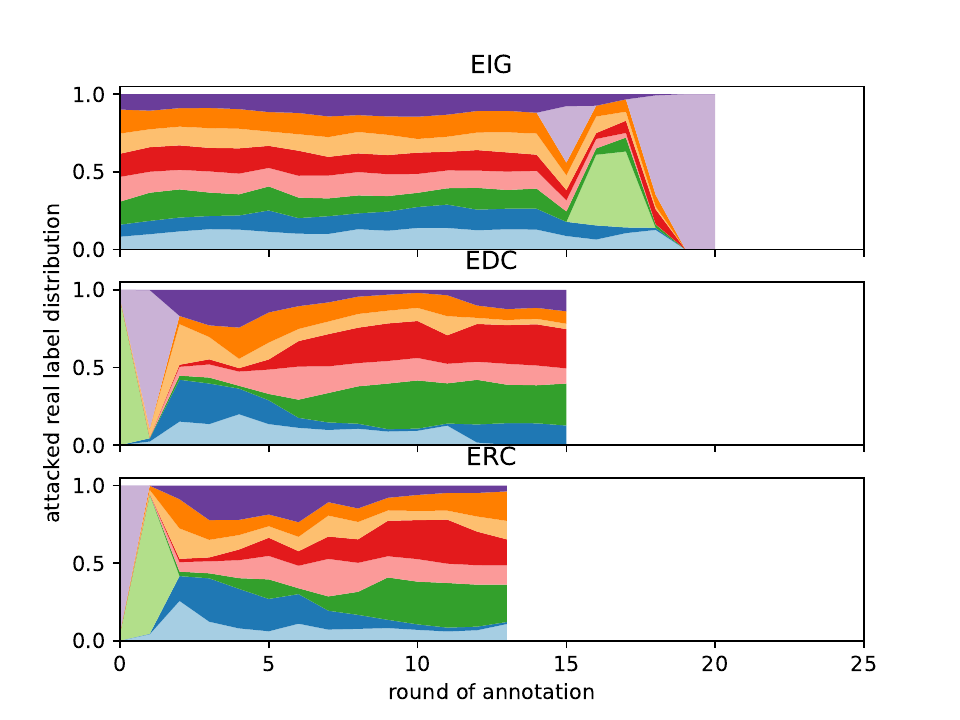}
    \caption{Adversarial CIFAR10}
  \end{subfigure}
  \caption{Label distribution among selected examples for CIFAR 10 (left) and adversarially perturbed CIFAR 10 (right). Light green and light purple mark the two classes made artificially easy.
}
  \label{fig:attack}
  \end{minipage}
\vspace{-10px}
\end{figure}

{\bf Sample uncertainty }
Classic uncertainty sampling
chooses data of high uncertainty.
This question is worth re-examining in the context of \prob{}.
To analyze the behavior of \prob{} learners
vis-a-vis uncertainty
we plot average prediction entropy of sampled data
for \prob{} learners
with different sampling strategies
(Figure~\ref{fig:entropy}).
Note that \prob{} learners using EIG
pick high-entropy data,
while \prob{} learners with EDC and ERC
choose examples with lower entropy predictions. 
The (perhaps) surprising performance of EDC and ERC
may owe to the cost structure of \prob{}.
While labels for examples with low-entropy predictions confer less information, they also come at lower cost.

{\bf Adversarial setting }
Because ERC goes after ``easy'' examples,
we test its behavior on a simulated dataset
where 2 of the {\it CIFAR10} classes (randomly chosen) are trivially easy.
We set all pixels white for one class
all pixels black for the other.
We plot the label distribution among the selected data over rounds of selection in
against that on the unperturbed {\it CIFAR10}
in Figure~\ref{fig:attack}.
As we can see, in the normal case,
EIG splits its budget among all classes
roughly evenly while EDC and ERC focus
more on different classes at different stages.
In the adversarial case,
EIG quickly learns the easy classes, thereafter focusing on the others until they are exhausted,
while EDC and ERC concentrate on exhausting the easy ones first.
Although EDC and ERC still manage to label all data
with less total cost than EIG,
this behavior might cost us when we have trivial classes, 
especially when our unlabeled dataset is enormous 
relative to our budget. 

\section{Related work}
\vspace{-2px}
\label{sec:related}
%
%


{\bf Binary identification:}
Efficiently finding answers with yes/no questions
is a classic problem~\citep{garey1972optimal} dubbed \emph{binary identification}.
\cite{hyafil1976constructing} proved
that finding the optimal strategy
given an arbitrary set of binary tests is NP-complete.
A well-known greedy algorithm called
{\it binary splitting} \citep{garey1974performance,loveland1985performance},
chooses questions that most evenly split the probability mass. 

{\bf  Active learning:}
Our work builds upon the AL framework
\citep{box1987empirical,cohn1996active,settles2010active} (vs. i.i.d labeling).
Classical AL methods select examples
for which the current predictor is most uncertain,
according to various notions of uncertainty:
\cite{dagan1995committee} selects examples
with \emph{maximum entropy} (ME)
predictive distributions,
while \cite{culotta2005reducing}
uses the \emph{least confidence} (LC) heuristic,
sorting examples in ascending order
by the probability assigned to the argmax.
\cite{settles2008active} notes
that annotation costs
may vary across data points
suggesting cost-aware sampling heuristics
but doesn't address the setting when costs change dynamically during training as a classifier grows stronger.
\cite{luo2013latent} incorporates structure among outputs into an active learning scheme in the context of structured prediction.
\citet{mo2016learning} addresses hierarchical label structure in active learning interestingly in a setting where subclasses are \emph{easier} to learn.
Thus they query classes more fine-grained than the targets,
while we solicit feedback on more \emph{general} categories.

{\bf Deep Active Learning}
Deep Active Learning (DAL)
has recently emerged as an active research area.
\cite{wang2016cost} explores a scheme
that combines traditional heuristics
with pseudo-labeling.
\cite{gal2017deep} notes that the softmax outputs of neural networks do not capture epistemic uncertainty \citep{kendall2017uncertainties},
proposing instead to use Monte Carlo samples
from a dropout-regularized neural network
to produce uncertainty estimates.
DAL has demonstrated success on NLP tasks.
\cite{zhang2017active} explores
AL for sentiment classification,
proposing a new sampling heuristic,
choosing examples for which
the expected update to the word embeddings is largest.
Recently, \cite{shen2018deep}
matched state of the art performance
on named entity recognition,
using just 25\% of the training data.
\cite{kampffmeyer2016semantic} and \cite{kendall2015bayesian}
explore other measures of uncertainty
over neural network predictions.
%
%
%

{\bf Learning from partial labels}
Many papers on learning from partial labels
\citep{grandvalet2004learning, nguyen2008classification, cour2011learning}
assume that partial labels are given a priori and fixed. 
\cite{grandvalet2004learning}
formalizes the partial labeling problem
in the probabilistic framework
and proposes a minimum entropy based solution.
\cite{nguyen2008classification}
proposes an efficient algorithm
to learn classifiers from partial labels
within the max-margin framework.
\cite{cour2011learning} addresses
desirable properties of partial labels
that allow learning from them effectively.
While these papers assume
a fixed set of partial labels,
we {\em actively} solicit partial feedback.
This presents new algorithmic challenges:
(i) the partial labels for each data point changes across training rounds;
(ii) the partial labels result from active selection, which introduces bias;
and (iii) our problem setup
requires a sampling strategy
to choose questions.



\vspace{-2px}

\section{Conclusion}
\vspace{-2px}
\label{sec:conclusion}
Our experiments validate the active learning with partial feedback framework 
on large-scale classification benchmarks. 
The best among our proposed \prob{} learners   fully labels the data with 42\% fewer binary questions as compared 
to traditional active learners. 
Our diagnostic analysis suggests 
that in \prob{}, it's sometimes more efficient to start with ``easier'' examples that can be cheaply annotated
rather than with ``harder" data as often suggested by traditional active learning.



\clearpage
\newpage

\bibliographystyle{apalike}
\bibliography{refs}



\appendix

\onecolumn
\section{Warm-starting Plot}
\label{sec:warmstart}
\begin{figure}[h]
  \centering  
  \includegraphics[width=\linewidth]{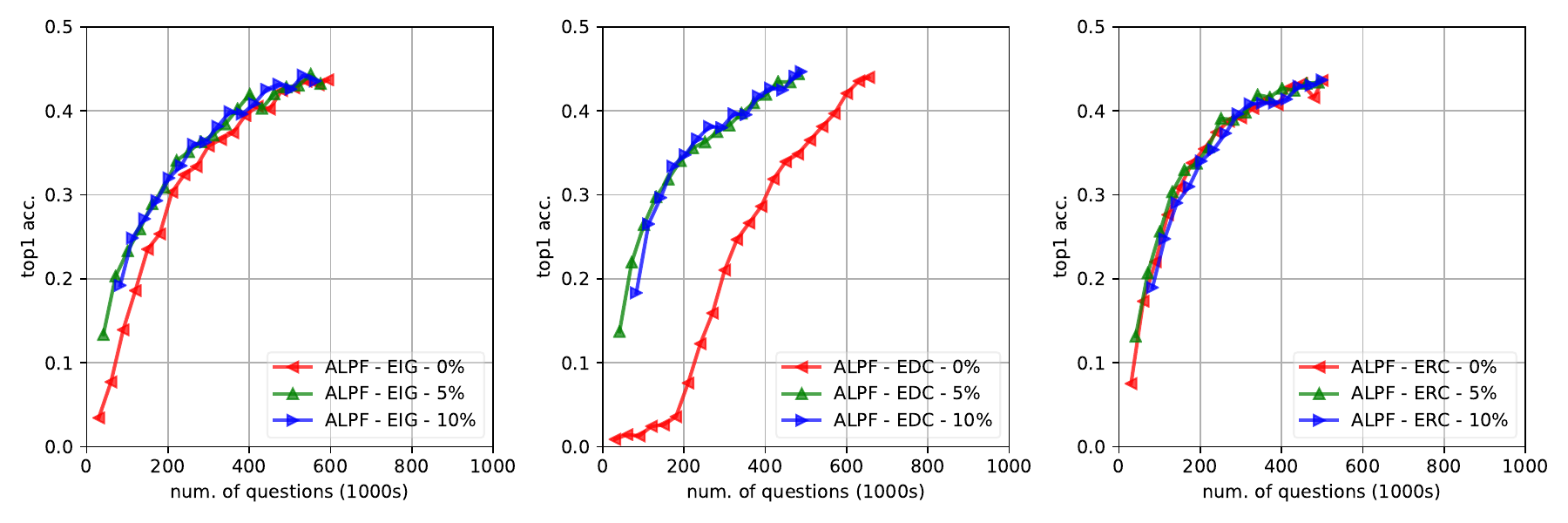}
  \caption{This plot compares our models under various amounts of warm-starting with pre-labeled i.i.d. data. We find that on the investigated datasets, ERC does benefit from warm-starting. However, absent warm-starting, EIG performs significantly worse and EDC suffers even more. We find that 5\% warmstarting helps these two models and that for both, increasing warm-starting from 5\% up to 10\% does not lead to further improvements.}
    \label{fig:warmstart}
\end{figure}

Figure~\ref{fig:warmstart} compares our strategies under various amounts of warm-starting with pre-labeled i.i.d. data.

\section{Updating models more frequently}
\label{sec:retrain}
\begin{figure}[ht]
  \vspace{-3em}
  \begin{center}
    Baseline \\
    \includegraphics[width=.32\linewidth]{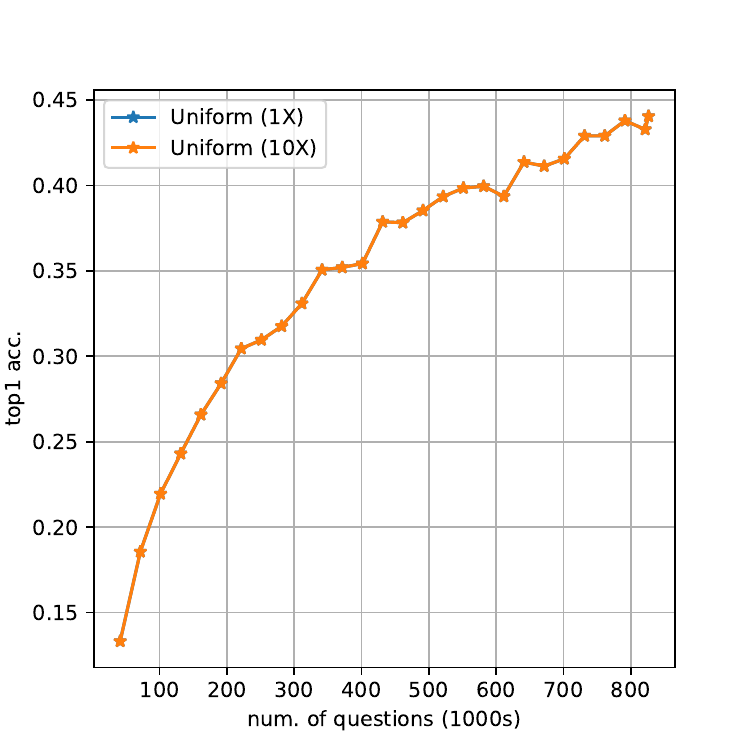}\\[1em]
    AD - (ME, LC)\\
    \includegraphics[width=.32\linewidth]{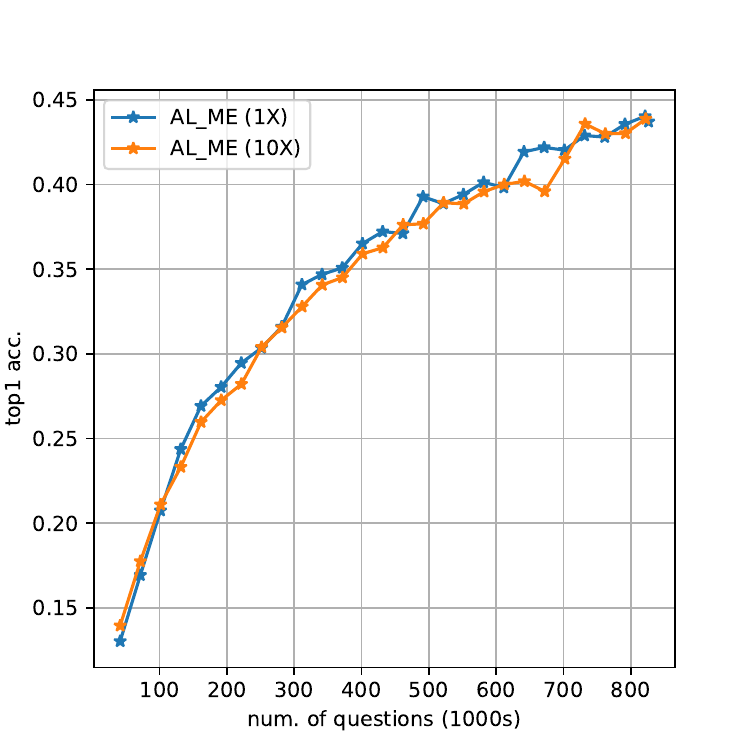}
    \includegraphics[width=.32\linewidth]{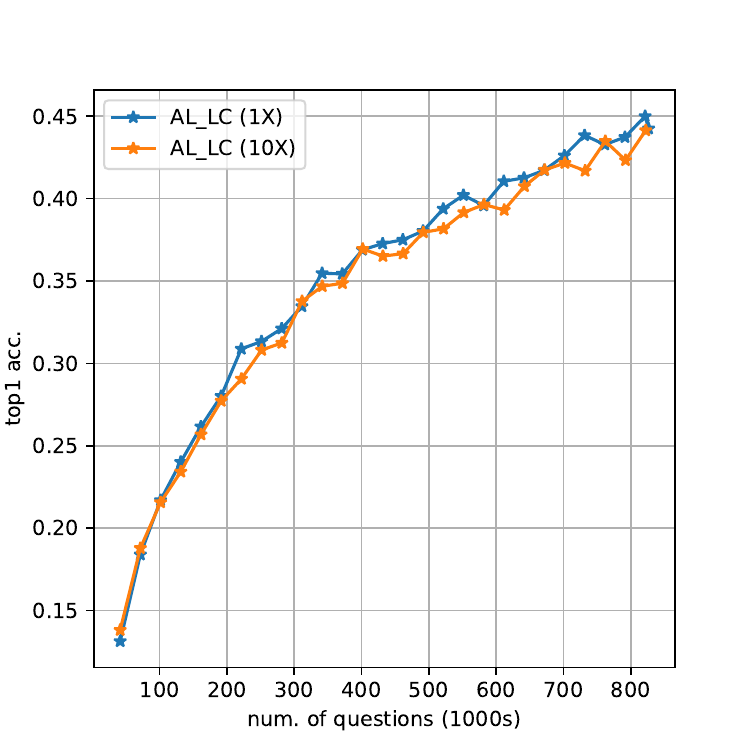}\\[1em]
    AQ - (EIG, EDC, ERC) \\
    \includegraphics[width=.32\linewidth]{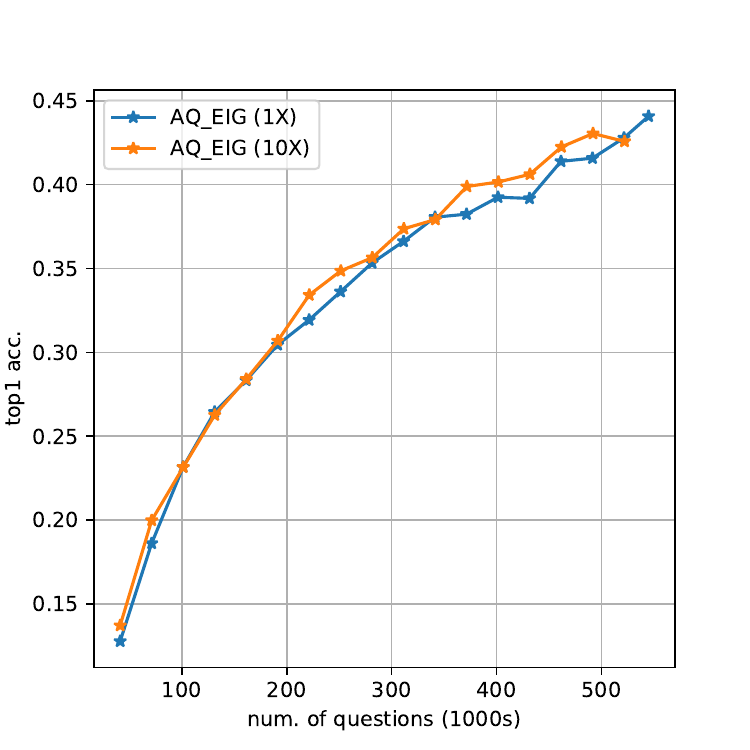}
    \includegraphics[width=.32\linewidth]{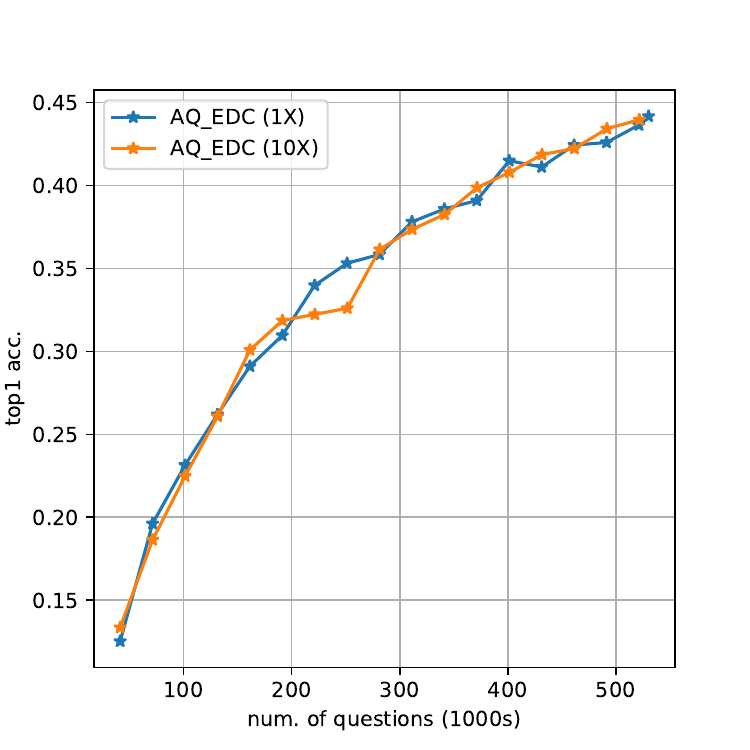}
    \includegraphics[width=.32\linewidth]{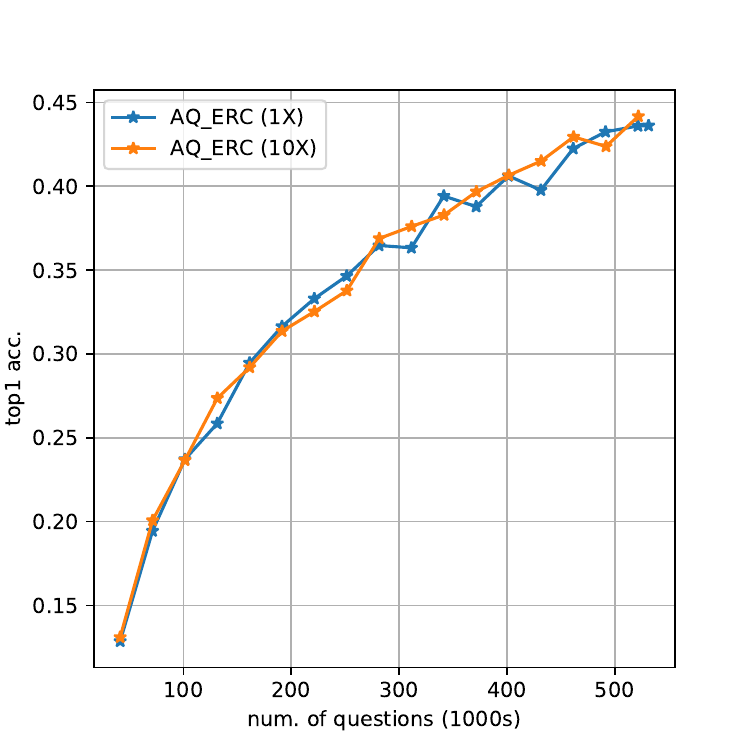}\\[1em]
    ALPF - (EIG, EDC, ERC)\\
    \includegraphics[width=.32\linewidth]{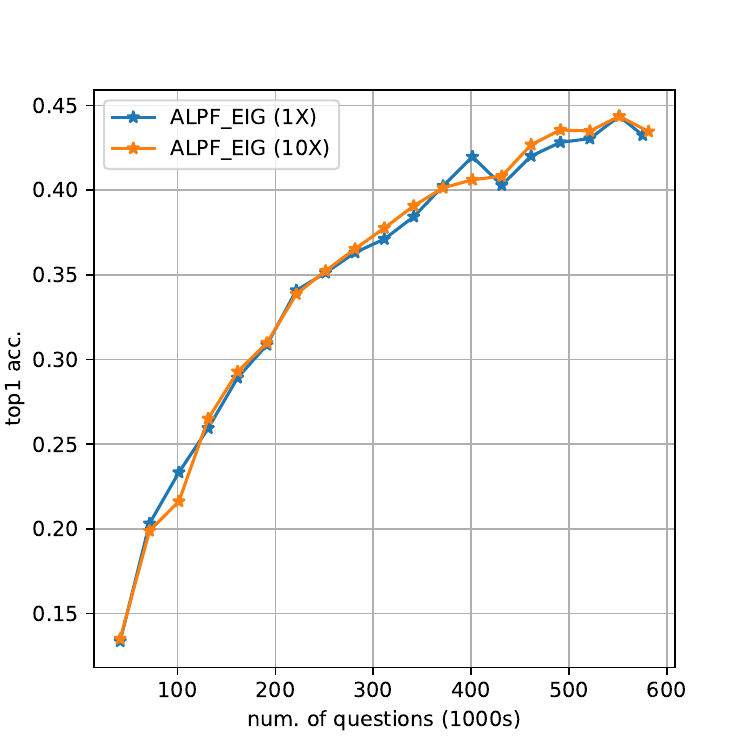}
    \includegraphics[width=.32\linewidth]{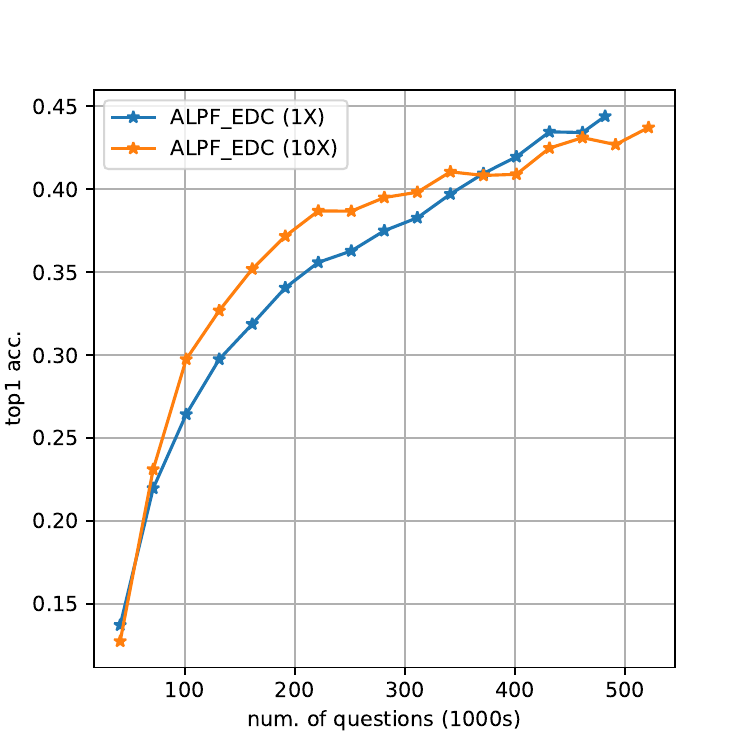}
    \includegraphics[width=.32\linewidth]{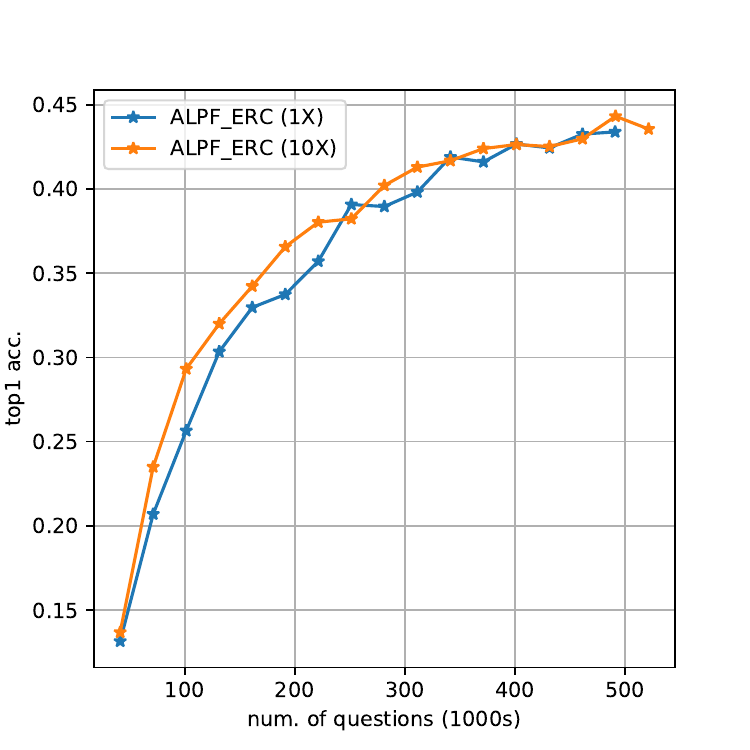}
    \caption{Updating models after every 30K questions (1X) vs. after every 3K (10X)}
    \label{fig:retrain}
  \end{center}
\end{figure}

On Tiny ImageNet, we normally re-initialize and train models from scratch for 75 epochs after every 30K questions. Since we found re-initialization is crucial for good performance, to ensure a fair comparison, we keep the same re-initialization frequency (i.e. every 30K questions) while updating the model by fine-tuning 5 epochs after every 3K questions. This results in 10X faster model updating frequency. As in Figure~\ref{fig:retrain} and Table~\ref{tab:retrain}, results show only ALPF-EDC and ALPF-ERC seem to benefit from updating 10 times more frequently

\begin{table}[t]
  \caption{Updating models after every 30K questions (1X) vs. after every 3K (10X)}
  \label{tab:retrain}
  \centering
  \begin{tabular}{l|cccccc|c}
    \toprule
    & \multicolumn{6}{c|}{\textbf{Annotation Budget}} & \textbf{Labeling Cost} \\
    \midrule
    & 10\%           & 20\%           & 30\%           & 40\%           & 50\%           & 100\% &      \\
    \midrule
    \multicolumn{8}{c}{\textbf{TinyImageNet}}\\
    \midrule
    \midrule
    Baseline            & 0.186          & 0.266          & 0.310          & 0.351          & 0.354          & 0.441 & 827k \\
    \midrule
    \midrule
    AL - ME             & 0.169          & 0.269          & 0.303          & 0.347          & 0.365          & -     & 827k \\
    AL - LC             & 0.184          & 0.262          & 0.313          & 0.355          & 0.369          & -     & 827k \\
    \midrule
    AL - ME - 10X& 0.177          & 0.260          & 0.304          & 0.341          & 0.359          & -     &  \\
    AL - LC - 10X& 0.188          & 0.257          & 0.308          & 0.347          & 0.369          & -     &  \\
    \midrule
    \midrule
    AQ - EIG            & 0.186          & 0.283          & 0.336          & 0.381          & 0.393          & -     & 545k \\
    AQ - EDC            & 0.196          & 0.291          & 0.353          & 0.386          & 0.415          & -     & 530k \\
    AQ - ERC            & 0.194          & 0.295          & 0.346          & 0.394          & 0.406          & -     & 531k \\
    \midrule
    AQ - EIG - 10X& 0.200          & 0.284          & 0.349          & 0.379          & 0.402          & -     & 522k \\
    AQ - EDC - 10X& 0.186          & 0.294          & 0.326          & 0.383          & 0.408          & -     & 522k \\
    AQ - ERC - 10X& 0.201          & 0.292          & 0.338          & 0.383          & 0.407          & -     & 522k \\
    \midrule
    \midrule
    ALPF - EIG          & 0.203          & 0.289          & 0.351          & 0.384          & 0.420          & -     & 575k \\
    ALPF - EDC          & \textbf{0.220} & 0.319          & 0.363          & 0.397          & 0.420          & -     & 482k \\
    ALPF - ERC          & 0.207          & \textbf{0.330} & \textbf{0.391} & \textbf{0.419} & \textbf{0.427} & -     & 491k \\
    \midrule
    ALPF - EIG - 10X& 0.199 & 0.293 & 0.352 & 0.391 & 0.406 & - & 581k \\
    ALPF - EDC - 10X& 0.231 & 0.352 & 0.387 & 0.410 & 0.409 & - & 521k \\
    ALPF - ERC - 10X& 0.235 & 0.342 & 0.382 & 0.417 & 0.426 & - & 521k \\
    \bottomrule
    \end{tabular}
  \vspace{-10px}
\end{table}




\end{document}